\documentclass[10pt,twocolumn,letterpaper]{article}

\usepackage{times}
\usepackage{epsfig}
\usepackage{graphicx}
\usepackage{amsmath}
\usepackage{amssymb}
\usepackage[utf8]{inputenc} 
\usepackage[T1]{fontenc}    
\usepackage{url}            
\usepackage{booktabs}       
\usepackage{amsfonts}       
\usepackage{nicefrac}  
\usepackage{microtype} 
\usepackage{bm}
\usepackage{bbm}
\usepackage{times}
\usepackage{authblk}
\usepackage{epsfig}
\usepackage{algorithm}
\usepackage{algpseudocode}
\usepackage{xcolor,colortbl}
\usepackage{graphicx}
\usepackage{color}
\usepackage{stmaryrd}
\usepackage{graphics}
\usepackage{mathptmx}
\usepackage{xspace}
\usepackage{dsfont}
\usepackage{mathtools}
\usepackage{tabularx, multirow}
\usepackage{gensymb}
\usepackage{indentfirst}

\usepackage[pagebackref=true,breaklinks=true,letterpaper=true,colorlinks,bookmarks=false]{hyperref}

\begin{document}

\title{Tree-gated Deep Regressor Ensemble For Face Alignment In The Wild}

\author[1]{Est\`ephe Arnaud}
\author[2]{Arnaud Dapogny}
\author[1, 2]{K\'evin Bailly}

\affil[1]{\small Sorbonne Universit\'e, CNRS, Institut des Systèmes Intelligents et de Robotique, ISIR, F-75005 Paris, France}
\affil[2]{\small Datakalab, Paris, France}

\date{}
\maketitle

\begin{abstract}
  Face alignment consists in aligning a shape model on a face in an image. It is an active domain in computer vision as it is a preprocessing for applications like facial expression recognition, face recognition and tracking, face animation, etc. Current state-of-the-art methods already perform well on "easy" datasets, \textit{i.e.} those that present moderate variations in head pose, expression, illumination or partial occlusions, but may not be robust to "in-the-wild" data. In this paper, we address this problem by using an ensemble of deep regressors instead of a single large regressor. Furthermore, instead of averaging the outputs of each regressor, we propose an adaptive weighting scheme that uses a tree-structured gate. Experiments on several challenging face datasets demonstrate that our approach outperforms the state-of-the-art methods.
\end{abstract}

\section{Introduction}
Face alignment consists in locating a set of landmarks on a face image (lips and eyes corners, pupils, nose tip). This is an important area of computer vision research, as it is an essential preprocess for applications such as face recognition, tracking, expression analysis or face synthesis.

Recently, regression-based methods appeared among the most successful approaches, providing impressive results on images with reasonable sources of variations in pose, illumination, expression or partial occlusions. These methods tackle the face alignment problem by directly learning the mapping between shape-indexed image appearances and the landmark locations. The regression is often performed in a cascaded manner: starting from an initial guess, a first regression step predicts a displacement for every landmark. This prediction is progressively refined by successive regressors that are trained to improve the predictions of the previous steps. In such a coarse-to-fine strategy, the first steps of the cascade usually capture large deformations, while the last steps focus on more subtle variations. The seminal work of Xiong et al. \cite{Xiong2013} propose to use successive linear regressions based on SIFT descriptors extracted around each landmark at the current position of the model. In the same vein, Ren et al. \cite{Ren2014} propose to learn shape-indexed pixel intensity differences with random forests instead of using SIFTs, in order to accelerate the feature extraction.

Deep learning techniques have been investigated to tackle the face alignment problem. For example, in \cite{Sun2013} the cascaded regressions are performed by stacking several deep convolutional networks (CNNs) in order to learn features in an end-to-end system instead of learning a regressor from handcrafted features. The Mnemonic Descent Method \cite{Trigeorgis2016} improves the feature extraction process by sharing the convolutional layer among all steps of the cascade that are performed through a recurrent neural network. This results in memory footprint reduction as well as better representation learning and a more optimized landmark trajectory throughout the different cascade steps.

\begin{figure}[!t]
\centering
\includegraphics[width=\linewidth]{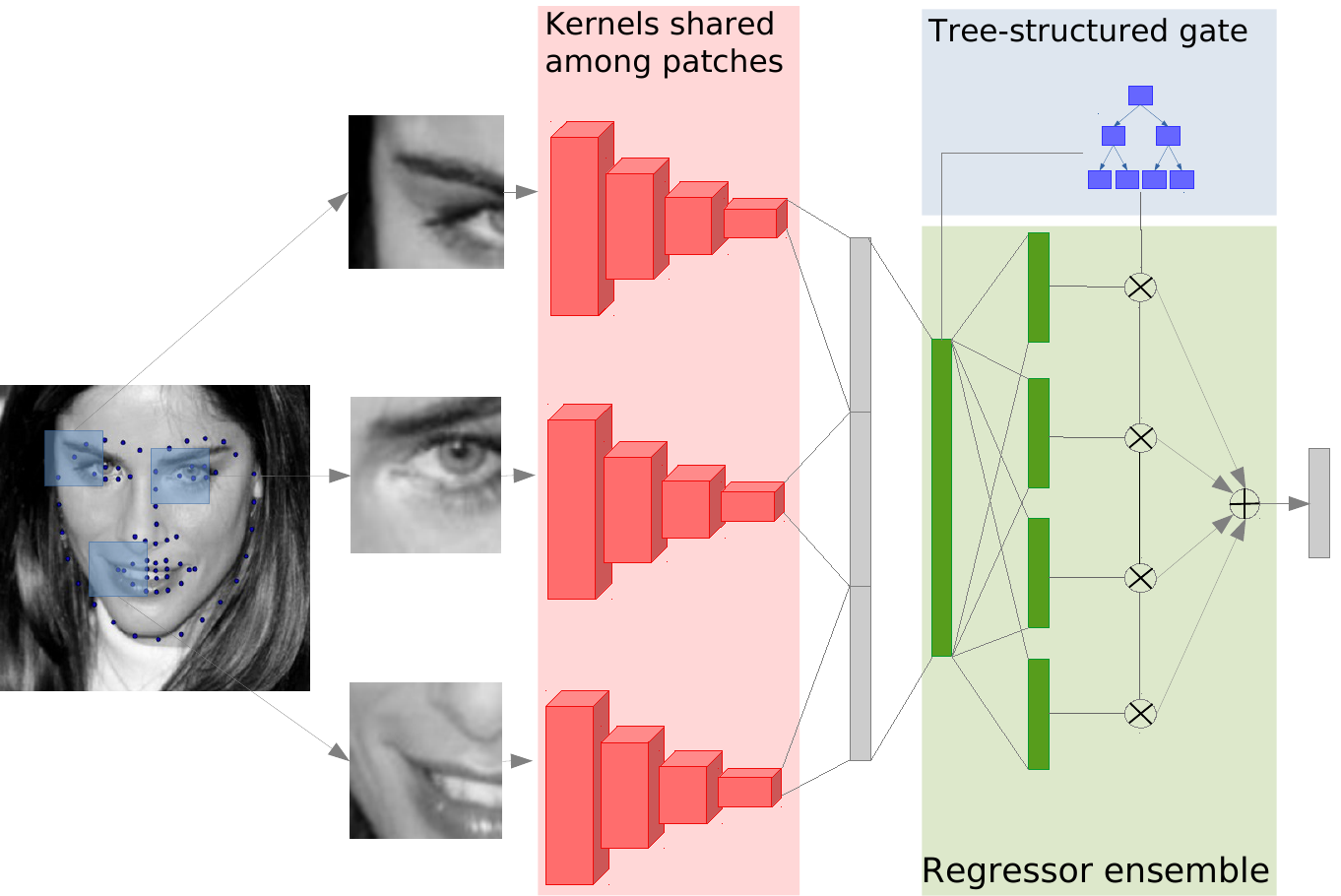}
\caption{Architecture of the proposed method. For each cascade stage, patches are extracted around feature points and a number of shared convolution kernels are applied to each patch. The output features are then flattened, concatenated and used as input of a regressor ensemble layer that predicts a displacement update for each landmark coordinate. The output of each regressor is then weighted by a tree-structured gate.}
\label{multi-mlp}
\end{figure}

Despite the success of cascaded methods, these are very sensitive to extreme conditions, where there are large head pose variations, expression variations, deformations, illumination variations, or occlusions induced by objects in front of the face (e.g. glasses, hands, hairs…). The appearance can then drastically change and corrupt the input features fed to the displacement regressor in the first steps of the cascades, causing errors that will be hard to overcome later on.
In order to address this limitation, ensemble methods can be used while keeping the advantages of a deep architecture. These methods improve the diversity of possible predictions by using an ensemble of independant regressors instead of a single large regressor, which leads to increased overall robustness. Since the optimization problem posed is not convex, a particular initialization for one regressor may lead to a different solution, and thus ensures diversity in the possible predictions while reducing the variance of the final response by averaging the responses of each predictor. Furthermore, instead of using a simple arithmetic mean to combine the regressors, we can learn an adaptative combination by using differentiable gating operators. In addition, the convolutional layers, the regressor ensemble and gating operators can be trained jointly in an end-to-end manner. Figure \ref{multi-mlp} illustrates this architecture. The contributions of this paper are thus three-folds:
\begin{itemize}
\item We show that using a regressor ensemble instead of a single large regressor is beneficial to the overall robustness of face alignment methods, particularly in the case where head pose variations and partial occlusion happen.
\item We propose an adaptative weighting scheme for combining the regressors, which uses tree-structured gates to learn a set of hierarchically clustered regressors. The tree-structured gates are learned as neural trees \cite{Kontschieder2016} to allow joint end-to-end training of regressor ensembles and gating operators.
\item We propose a complete face alignment system that runs in real-time and outperforms state-of-the-art approaches on several databases.
\end{itemize}

%%%%%%%%%%%%%%%%%%%%%%%%%%%%%%%%%%%%%%%%%%%%%%%%%%%%%%%%%%%%%%%%%%%%%%%%%%%%%%%%
\section{Related work}
We will first present some face alignment methods that are robust to different sources of variation, then the different studies that have been carried out on the unification of ensemble methods with deep learning techniques.

Most robust face alignment methods fall into two categories: first, some models aim to address robustness to only one source of variation. The architecture and training of it are then designed to specifically address this specific problem. For example, some models are specialized on occlusion handling and explicitly predict the occluded part of the face \cite{Burgos-Artizzu2013, Ghiasi2014, Yu2014, zhang2016occlusion}. However, several limitations must be taken into account: learning such models is most of the time supervised, and therefore requires labelled data in sufficient quantity (e.g. in terms of occluded face part), which is a major requirement. Furthermore, especially in the case of occlusion, it is very hard and time consuming to gather a dataset that accounts for all possible occlusion configurations. Additionnally, focusing on a single source of variation (e.g. occlusion) do not generalize well to other kinds of variations (e.g. non-frontal head pose variation).

Other methods aim to be robust to all sources of variation. Wu et al. \cite{Wu2017} proposes a global framework, trained in a cascaded manner, which simultaneously performs facial landmarks localization, occlusion detection, head pose estimation and deformation estimation with separate modules. Relationships between these allows the modules to benefit from each other. However, once again, each module requires additional annotations in the trainset (e.g. occlusion and head pose labelling). Zhang et al. \cite{zhang2016learning} improves the performance of deep learning based approaches by learning multiple tasks simultaneously thanks to auxiliary attributes. Training is then better conditioned and allows to increase the capacity of generalization. For example, knowing if glasses are present on the face can improve the model's robustness to occlusions. Honari et al. \cite{honari2018improving} also use auxiliary attributes for landmarks localization, but improve learning with a semi-supervised procedure. However, these techniques require additional data, either unlabelled or annotated with auxiliary attributes.

Our approach falls into the second category: it aims to be robust to all sources of variation present in the trainset, but without requiring additional annotations other than the location of landmarks. To do so, we combine the advantages of ensemble methods with those of deep learning techniques, a technique which has already been explored in the deep learning literature:
a first approach for combining deep learning and ensemble methods is to craft a differentiable ensemble architecture in order to allow end-to-end parameter learning. Kontshieder et al. \cite{Kontschieder2016} design a differentiable deep neural forest, by unifying the divide-and-conquer principle of decision trees (allowing to cluster data hierarchically) with the representation learning from deep convolution networks. Each predictor is a binary tree, whose split nodes contain routing functions, defining the probability of reaching one of the sub-trees. The probabilistic routing functions are differentiable, allowing these neural trees to be integrated into a fully-differentiable system. The forest corresponds to a set of trees, whose final output corresponds to the simple average of the outputs of each tree.
Since then, other models have sought to generalize neural forest \cite{Tanno2018}\cite{Ioannou2016}, by integrating upstream convolutionnal or multi-layer perceptron layers within routing functions to learn more complex input partitionings, leading to higher performance.
Dapogny et al. \cite{Dapogny2018a} used neural forests for face alignment, with promising results. However, their model uses handcrafted features by using SIFT descriptors. In addition, to adapt neural forest to the case of regression, they use neural trees whose leaves are fixed and correspond to a Gaussian draw on the remaining displacements from the training data (with little variations). Their model then seeks to optimally combine fixed leaves. This training procedure involving fixed leaves can theoretically lead to rigid responses, reducing the degree of freedom in possible predictions.

A second approach may be to parallelize a set of small networks instead of a single large network. The idea of using a set of regressors within an end-to-end system was firstly introduced by Jacobs et al. \cite{Jacobs1991} and more recently taken up by Eigen et al. \cite{eigen2013}, presenting promising results and well adapted to our problem. Eigen et al. design a Mixture-of-Experts (MoE) layer, consisting in jointly learning a set of ``expert'' subnetworks with gates, allowing to learn to combine a number of experts depending on the input. In the same vein, Shazeer et al. \cite{shazeer2017outrageously} introduce sparsity in MoE in order to save computation and to increase representation capacity.

%%%%%%%%%%%%%%%%%%%%%%%%%%%%%%%%%%%%%%%%%%%%%%%%%%%%%%%%%%%%%%%%%%%%%%%%%%%%%%%%
\section{Framework overview}

\begin{figure*}[!ht]
	\centering
	\includegraphics[width=\linewidth]{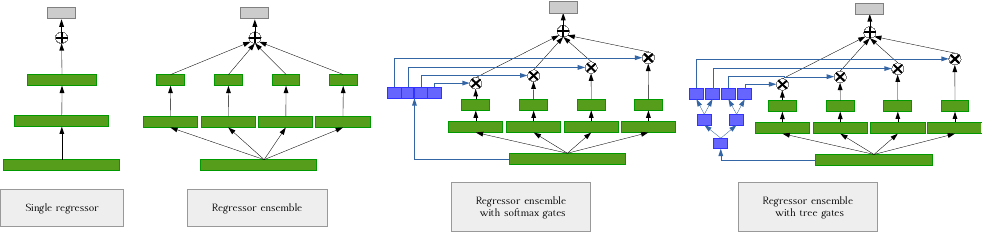}
	\caption{Overview of regressor ensemble layer. Regressors and gating operators are depticted in green and blue, respectively. \textbf{No gates}: the output (gray box) is the arithmetic mean of $T$ independant regressors. \textbf{Softmax gates}: the regressors are weighted by the output of a $T$-dimensional softmax layer. \textbf{Tree gates}: the regressors are weighted by the terminal (leaf) probabilities of a $\log_2(T)$-deep neural tree.}
	\label{architectures}
\end{figure*}

We propose a cascaded shape regression framework based on deep regressor ensemble for face alignment. At each stage of the cascade, CNNs are used to extract features around each landmark current location estimation, with weights sharing between landmarks to reduce the memory footprint. Then, a set of weak regressors estimates the mapping between a concatenation of those features and a landmark displacement. Finally, we explore the impact of using gates or not, and define a new type of gate in order to learn a set of hierarchically clustered regressors.

\subsection{Cascaded shape model}
For an image $\mathcal{I}$, we start from the mean shape $\textbf{s}^0 \in \mathbb{R}^{136}$ and learn a cascade of mappings $\textbf{s}^0 + \sum_k \delta \textbf{s}^k$ between shape-indexed features $\phi(\textbf{s}^k,\mathcal{I})$ and displacements $\delta \textbf{s}^k(\phi(\textbf{s}^k,\mathcal{I}))$ in shape space. More specifically, at each stage $k$ of the cascade, we extract features at the current feature point locations $\textbf{s}^k$ and train a regressor to learn the mapping between those features and the displacement. In what follows, we present details on our deep architecture for extracting shape-indexed features and training the regressors.

\subsection{Feature extraction}
At it is ubiquitous among recent face alignment methods, we use convolutional neural nets for feature extraction. To limit the number of parameters we share the convolution kernels between the different patches. For each cascade stage $k$, we use a sequence of $4$ strided convolution layers (2 $5 \times 5$ with followed by 2 $3 \times 3$ kernels) followed by $1$ convolution ($1 \times 1$ kernel) with ReLU activation to process the $68$ patches extracted around each landmark. We use $20 \rightarrow 40 \rightarrow 80 \rightarrow 160 \rightarrow 30$ feature channels, which makes $30$ $2 \times 2$ features maps for each landmark before flattening. Thus, the output concatenated and flattened feature vector $\phi(\textbf{s}^k,\mathcal{I})$ is of size $8160$. A fully-connected layer is then applied and the final vector is of size $2048$.

\subsection{Regression layer}

We now aim at regressing the displacement $\delta \textbf{s}^k$ from the extracted feature vector $x=\phi(\textbf{s}^k,\mathcal{I})$. Figure \ref{architectures} shows a number of interesting configurations that can be employed for that matter.

\subsubsection{Single regressor}

A basic approach consists in learning a single multi-layer perceptron (single regressor or SR) with ReLU activation as a regressor:

\begin{equation}\label{singlemlp}
\delta \textbf{s}^k = w_1.\max(0,w_0.x+b_0)+b_1
\end{equation}

\subsubsection{Regressor ensemble}
An alternative approach consists in using a simple combination of small multi-layer perceptrons. Each of these is composed of a hidden layer with ReLU activation and an affine output layer:

\begin{equation}\label{ensemblemlp}
R^{l}_{\Theta}(\phi(\textbf{s}^k,\mathcal{I})) = w_1^l.\max(0,w_0^l.x+b_0^l)+b_1^l
\end{equation}

With $\Theta = \{w_0^l, b_0^l, w_1^l, b_1^l\}$. As it is classifcal in the frame of ensemble methods, the displacement is then predicted as an average of $L$ independant ``weak'' regressors:

\begin{equation}\label{nogate}
\delta \textbf{s}^k = \frac{1}{L}\Sigma_lR^l_{\Theta}(\phi(\textbf{s}^k,\mathcal{I}))
\end{equation}

In what follows, we refer to this model as the regressor ensemble (RE). From an ensemble learning standpoint, combining several independant weak regressors allows to output more robust predictions.

\subsubsection{Simple gated ensemble}

The aforementionned architecture is similar to the mixture of experts (MoE) layer introduced in \cite{shazeer2017outrageously}. However, \cite{shazeer2017outrageously} use a softmax gating function to select the relevant blocks and (possibly) increase computational efficiency. Formally:

\begin{equation}\label{nogate}
\delta \textbf{s}^k = \Sigma_l G_{Smax}^l(\phi(\textbf{s}^k,\mathcal{I})) R^l_{\Theta}(\phi(\textbf{s}^k,\mathcal{I}))
\end{equation}

With $G_{Smax}(\phi(\textbf{s}^k,\mathcal{I})) = softmax(\phi(\textbf{s}^k,\mathcal{I}))$ denotes the gating function that weights the $L$ experts and $G_{Smax}^l(\phi(\textbf{s}^k,\mathcal{I}))$ is the $l^{th}$ coordinate of this vector. Compared to the simple regressor ensemble layer, this softmax-gated regressor ensemble (Soft-GRE) allows to learn an adaptative combination of regressors.

\subsubsection{Tree-gated ensemble}

In order to learn a more potent gating function we use a single neural tree. A neural tree \cite{Kontschieder2016} is composed of subsequent soft, probabilistic routing functions $d_n$, that represents the probability to reach the left child of node $n$. Formally, we have:
\begin{equation}
\begin{array}{ccccc}
d_n(.) & : & \mathcal{X} & \to & [0, 1] \\
& & x & \mapsto & \sigma(w_n.x + b_n) \\
\end{array}
\end{equation}
Thus, $d_n$ is defined as a single neuron that takes as input the feature vector $x$, with parameters $\{w_n, b_n\}_{n \in \mathcal{N}}$ and $\sigma$ a sigmoid activation.

For an example $x$, the probability $\mu_l$ to reach a leaf $l \in \{0,...,L-1\}$ is computed as a product of the successive activations $d_n$ down the whole tree. Therefore:
\begin{equation}
\mu_l(x) = \prod_{n \in \mathcal{N}}d_n(x)^{\mathds{1}_{l \swarrow n}}(1 - d_n(x,))^{\mathds{1}_{l \searrow n}}
\end{equation}
where $l \swarrow n$ is true if $l$ belongs to the left subtree of node $n$, and $l \searrow n$ is true if $l$ belongs to the right subtree.

We define our tree-structured gate $G_{Tree}(\phi(\textbf{s}^k,\mathcal{I})) = (\mu_0(\phi(\textbf{s}^k,\mathcal{I})), ..., \mu_{L-1}(\phi(\textbf{s}^k,\mathcal{I})))$ as the concatenation of the $2^\mathcal{D}$ leaves probabilities of a single neural tree of depth $\mathcal{D}$, that takes as input the feature vector $\phi(\textbf{s}^k,\mathcal{I})$. Once again, the displacement is obtained by weighting each regressor $l$ by the $l^{th}$ gate coordinate:
\begin{equation}\label{nogate}
\delta \textbf{s}^k = \Sigma_l G_{Tree}^l(\phi(\textbf{s}^k,\mathcal{I})) R^l_{\Theta}(\phi(\textbf{s}^k,\mathcal{I}))
\end{equation}

This tree-gated weighting scheme (Tree-GRE) allows to learn a more efficient, hierarchical regressor weighting, as compared to softmax gating.

\subsection{Implementation details}

Training is done on 150x150-grayscale images with $32 \times 32$ patches centered around the 68 landmarks. The pixels are centered so that they take values in $[-1, 1]$. We train $4$ cascade stages and apply data augmentation as it is common in the literature: for each image we augment the initial mean shape by a random translation factor $t \sim \mathcal{N}(0, 10)$ and a random scaling factor $s \sim \mathcal{N}(1, 0.1)$. Optimization is applied \textit{via} ADAM optimizer \cite{Sharma2017} with a learning rate of 0.001.

\section{Experiments}

In what follows, we perform extensive architecture and hyperparameter ablation study to show the interest of using ensemble of regressors as well as gates.

\subsection{Datasets}

The 300W dataset was introduced by the I-BUG team \cite{Sagonas2015}\cite{menpo14} and is considered the benchmark dataset for training and testing face alignment models, with moderate variations in head pose, facial expressions and illuminations. It also embraces a few occluded images. The 300W dataset consists in four datasets: \textbf{LFPW} (811 images for train / 224 images for test), \textbf{HELEN} (2000 images for train / 330 images for test), \textbf{AFW} (337 images for train) and \textbf{IBUG} (135 images for test), for a total of 3148 images annotated with 68 landmarks for training the models. For comparison with state-of-the art methods, we refer to LFPW and HELEN test sets as the \textit{common} subset and I-BUG as the \textit{challenging} subset of 300W, as it is commonly done in the literature.

The \textbf{COFW} or Caltech Occluded Faces in the Wild dataset \cite{Burgos-Artizzu2013}, is an "in-the-wild" dataset containing only occluded data. It is a benchmark dataset to test the robustness of occlusion models. COFW contains 500 images for train and 507 images for test. The models are trained with 68 landmarks annotated for each image of the datasets. However, COFW only contains images with 29 annotated landmarks. Thus, we use the method proposed in \cite{Ghiasi2014} to perform a linear mapping between the predictions made on the 68 landmarks to the 29 landmarks, as it is a common practice on this dataset.

We use the average point-to-point distance between the ground truth and the predicted shape, normalized by the inter-pupil distance, as it is classically done in the literature. For clarity, in what follows, we omit the ’\%’ symbol.

\subsection{Model comparison}

\begin{table}[!h]
	\centering
	\caption{Comparison of different regressors ($\%$ error). The best and second best models are resp. highlighted in red and blue.}
	\begin{tabular}{c|c|c|c|c}
		\hline
		Database&SR&RE&Soft-GRE&Tree-GRE\\
		\hline
		\hline
		LFPW&3.98&3.82&\underline{\textcolor[rgb]{0,0,1}{3.78}}&\textbf{\textcolor[rgb]{1,0,0}{3.76}}\\
		HELEN&4.48&4.33&\underline{\textcolor[rgb]{0,0,1}{4.27}}&\textbf{\textcolor[rgb]{1,0,0}{4.24}}\\
		I-BUG&8.86&8.53&\underline{\textcolor[rgb]{0,0,1}{8.39}}&\textbf{\textcolor[rgb]{1,0,0}{8.31}}\\
		COFW&6.03&\underline{\textcolor[rgb]{0,0,1}{5.86}}&5.92&\textbf{\textcolor[rgb]{1,0,0}{5.75}}\\
		\hline
	\end{tabular}
	\label{ensemblecomp}
\end{table}

First, in order to compare the different architectures as accurately as possible, we adapt them so that we always have the same number of weights to learn. In particular, the different architectures proposed are as follows:
\begin{itemize}
\item \textbf{SR}: one single "large" regressor with one hidden layer $2048 \times 5120 \rightarrow 5120\times136$.
\item \textbf{RE}: an ensemble of 128 regressors with one hidden layer $2048\times40 \rightarrow 40\times136$.
\item \textbf{Soft-GRE}: an ensemble of 128 regressors with one hidden layer $2048\times40 \rightarrow 40\times136$ with a single 128-dimensional softmax layer.
\item \textbf{Tree-GRE}: an ensemble of 128 regressors with one hidden layer $2048\times40 \rightarrow 40\times136$, and a neural tree of depth 7 (with $2^7=128$ leaves).
\end{itemize}
With this configuration, each model has approximatly 114 million parameters total, for the 4 cascade steps and with convolutional/regression layers. The results are showed on Table \ref{ensemblecomp}. In particular, the non gated regressor ensemble (RE) is more robust than a single regressor (SR), especially on data presenting large variations in pose (I-BUG) or partial occlusions (COFW). Moreover, adding gates improves performance, especially with tree-structured gates. Softmax gates improve robustness to pose (I-BUG), but not to occlusions (COFW), contrary to tree-structured gates which improve performance for all configurations.

\subsection{Comparison with state-of-the-art approaches}

\begin{table}[!ht]
	\centering
		\caption{Comparison with state-of-the-art methods. Best and second best results are resp. highlighted in red and blue.}
		\begin{tabular}{c|c|c|c}
			\hline
			Method   & Common & Challenging & COFW \\
			\hline
			\hline
			RCPR \cite{Burgos-Artizzu2013} & 6.18  & 17.3 & 8.50\\
			SDM \cite{Xiong2013}           & 5.57  & 15.4 & 7.70\\
			PIFA \cite{jourabloo2017pose}  & 5.43 & 9.98 & -\\
			LBF \cite{Ren2014}             & 4.87  & 11.98 & 13.7\\
			TCDCN \cite{zhang2016learning} & 4.80 & 8.60 & -\\
			CSP-dGNF \cite{Dapogny2018a}   & 4.76  & 12.00  & - \\
			ERCLM \cite{Boddeti2017}       & 4.58  & 8.90 & - \\
			RAR \cite{xiao2016robust}      & \underline{\textcolor[rgb]{0,0,1}{4.12}} & 8.35 & \underline{\textcolor[rgb]{0,0,1}{6.03}} \\
			RCN$^+$ \cite{honari2018improving} & 4.20 & \textbf{\textcolor[rgb]{1,0,0}{7.78}} & -\\
			DRDA \cite{zhang2016occlusion} & - & 10.79 & 6.46\\
			SFLD \cite{Wu2017} & - & - & 6.40\\
			\hline
			\hline
			Tree-GRE & \textbf{\textcolor[rgb]{1,0,0}{4.05}}  & \underline{\textcolor[rgb]{0,0,1}{8.31}} & \textbf{\textcolor[rgb]{1,0,0}{5.75}} \\
			\hline
		\end{tabular}
	\label{res300w}
\end{table}

Table \ref{res300w} shows a comparison between our approach and other recent state-of-the-art methods on both 300W (common and challenging subsets) and COFW databases. Our model outperforms these approaches on both 300W-common and COFW databases. Only \cite{honari2018improving} outperforms our model on the challenging subset of 300W. This approach, however, leverages additional unsupervised data to regularize the learning. Our Tree-GRE model also sets a new state-of-the-art of COFW database, as the regressor ensembles and tree-gating scheme allow to substantially increase the overall robustness of the model, particularly in the case of partial occlusions.

Last but not least, our method is very fast as it operates at $4,72$ ms per image on a NVIDIA GTX 1080 GPU, and thus can run at 211 fps. Furthermore, In \cite{shazeer2017outrageously}, softmax-gated models are used to reduce the computational load by keeping only a small number (top-k) of experts. With tree-structured gates, an interesting direction would be to evaluate the tree gate in a greedy layerwise fashion as in \cite{Dapogny2018a}, and keep only the regressor corresponding to the maximum probability leaf to further reduce the computational cost.

\subsection{How do gated models behave?}

\begin{figure*}[!h]
	\centering
	\includegraphics[width=\linewidth]{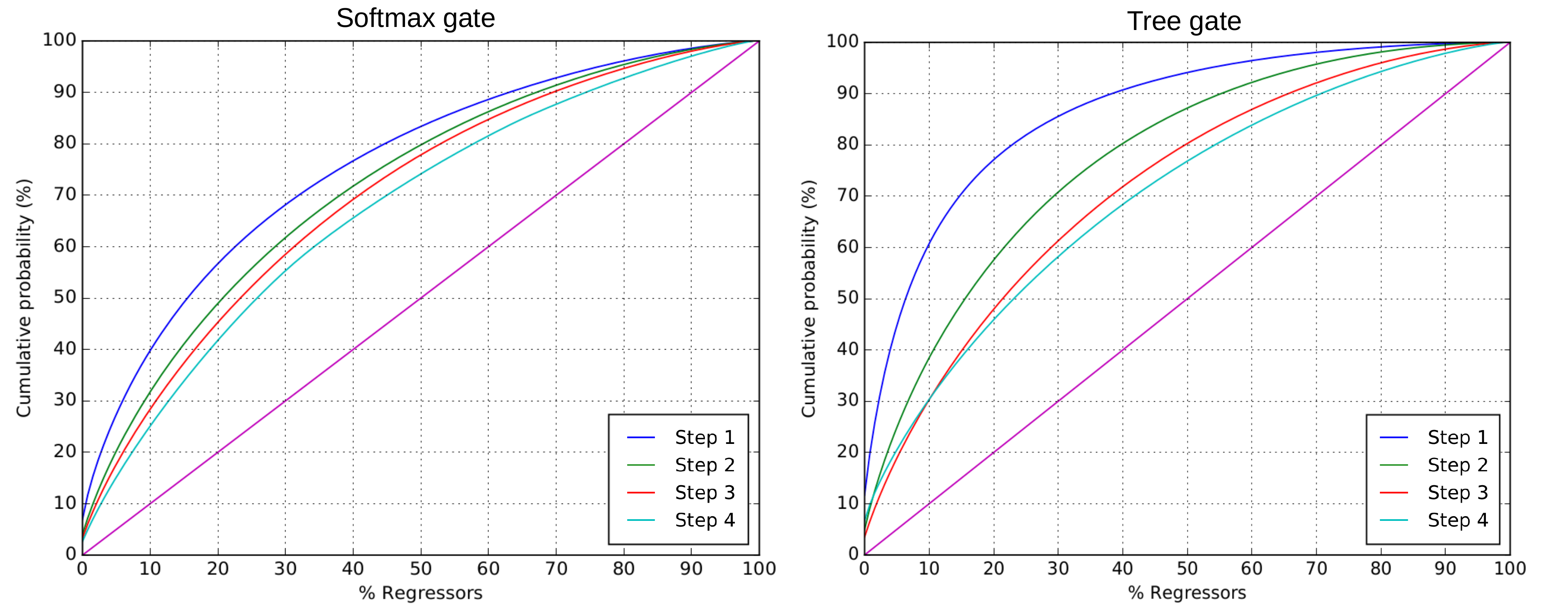}
	\caption{Cumulative top-scoring regressor distribution and comparison between softmax and tree gates.}
	\label{graphgate}
\end{figure*}

Next, we conduct some experiment to provide insight on how the gated models behave and the differences between softmax and tree-gated models. Figure \ref{graphgate} shows the average on HELEN testset of the cumulative sum of the gates probabilities of the regressors, sorted in descending order. For instance, softmax gates at cascade step 1 allow $10\%$ of the regressors to explain $40\%$ of the final prediction, while tree gates allow $10\%$ of the regressors to explain $60\%$ of the final prediction. Overall the first cascade steps necessitates less regressors than the subsequent steps, as the models only need to capture coarse landmark displacement corresponding to a translation of the initial shape, or to head pose variation. Later on, more regressors are needed to capture more subtle information adequately. This effect is more conspicuous for the tree-gated model than for the softmax-gated model, as the tree-gated model generally uses less regressors than the softmax-gated model. This mean that using a tree gate allows a better clustering of the regressors and an overall more efficient learning. Furthermore, the fact that tree-gating promotes the use of a more restricted number of regressors further justify the evaluation of a small fraction of the network, e.g. using greedy evaluation as in \cite{Dapogny2018a}.

\begin{figure*}[!ht]
	\centering
	\includegraphics[width=\linewidth]{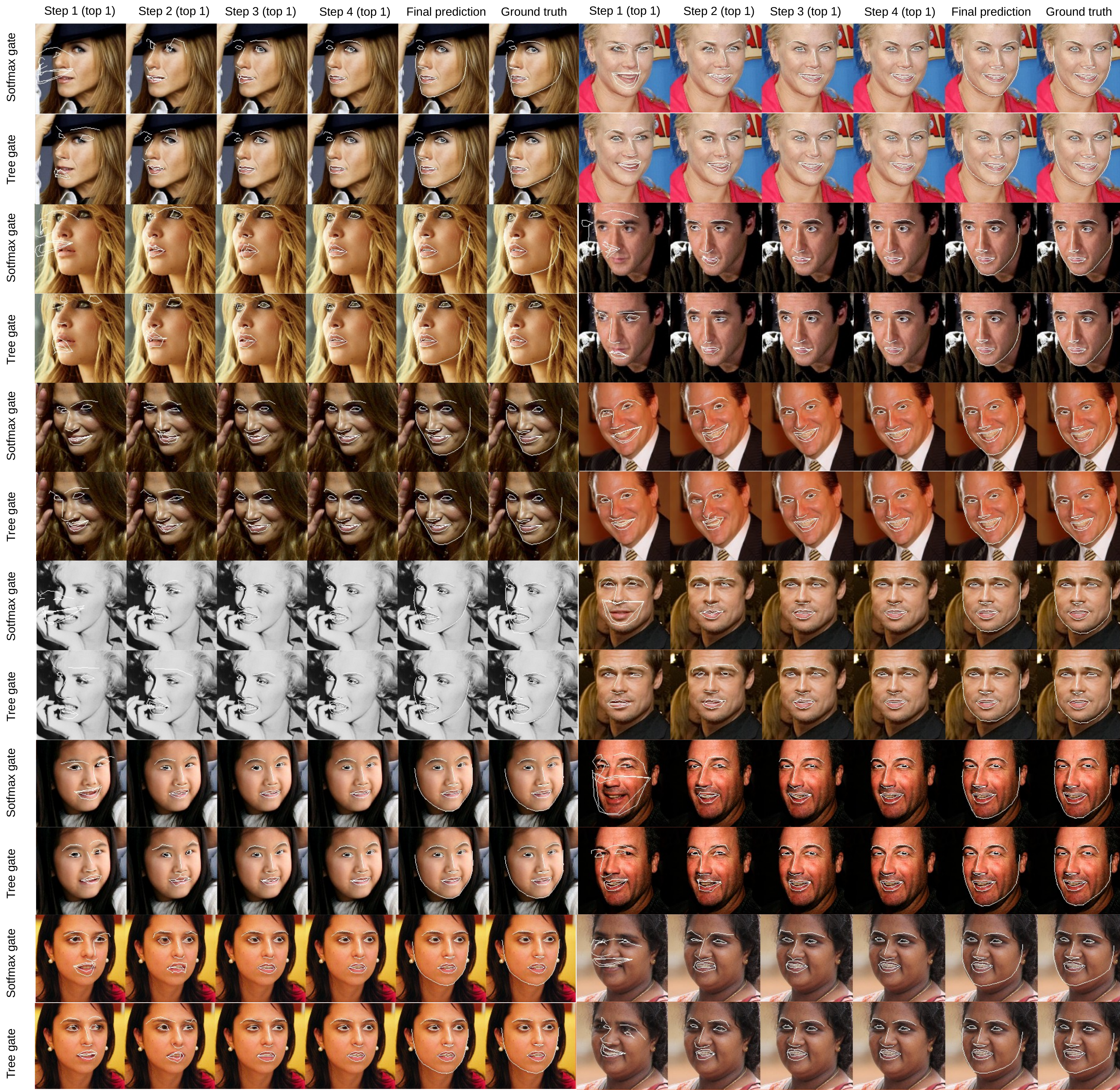}
	\caption{Visualisations of the predictions outputted for each cascade step with only the top (maximum value of either softmax or tree gate) regressor. Images from 300W testset.}
	\label{visus}
\end{figure*}

This phenomenon is also echoed by the qualitative visualisations in Figure \ref{visus}, which represents the alignment obtained using only the maximum probability regressor for various cascade steps for examples that present partial occlusions as well as head pose variations. The regressed updates are qualitatively a lot better with the tree-gated model, indicating that the tree gate allows to more efficiently select the most relevant regressor, and thus, at train time, learn an ensemble of better clustered experts. Also notice how the predicted shape matches the ground truth (two columns on the right), particularly in the case of the tree-gated model.

\section{Conclusion}
In this paper, we have presented a new architecture for face alignment robust to in-the-wild scenarios. The proposed architecture extracts features by sharing weights between all patches centered around each landmark allowing for better representation, as well as memory footprint reduction. Based on these features, it predicts a shape update using an ensemble of regressors to take advantage of deep learning techniques and ensemble methods. These regressors are combined in an adaptative way using a tree-structured gate. The regressor ensemble, the tree gate and the convolutional layers are optimized jointly, leading to a greater robustness to strong variations in pose or occlusions. We show empirically that we achieve state-of-the-art performance on multiple challenging databases while keeping the evaluation runtime low. Future direction involve using tree-gated ensemble layers to learn intermediate representations, as well as greedy evaluation of the soft decision trees to dramatically reduce the computational burden.

\section*{Acknowledgment}

This work has been supported by the French National Agency (ANR) in the frame of its Technological Research JCJC program (FacIL, project ANR-17-CE33-0002).

{
\bibliographystyle{ieee}
\bibliography{biblio}
}

\end{document}